\documentclass[letterpaper]{article} 
\usepackage[preprint]{aaai2027}  
\usepackage[hyphens]{url}  
\usepackage{graphicx} 
\usepackage{natbib}  
\usepackage{caption} 
\usepackage{algorithm}
\usepackage{algorithmic}

\usepackage{newfloat}
\usepackage{listings}
\DeclareCaptionStyle{ruled}{labelfont=normalfont,labelsep=colon,strut=off} 
\floatstyle{ruled}
\newfloat{listing}{tb}{lst}{}
\floatname{listing}{Listing}

\usepackage{booktabs}
\usepackage{colortbl}
\definecolor{ewhl}{HTML}{E6F3FA}

\usepackage{amsmath}
\usepackage{amssymb}
\newcommand{\cmark}{\ensuremath{\checkmark}}
\newcommand{\xmark}{\ensuremath{\times}}

\title{EndoWAM: A Grounded World-Action Model\\for Generalizable Endoscopic Navigation}
\author{
    Jinsong Lin\textsuperscript{\rm 1}\equalcontrib,
    Zikang Pan\textsuperscript{\rm 1}\equalcontrib,
    Wanhao Liu\textsuperscript{\rm 1}\equalcontrib,
    Chi Kit Ng\textsuperscript{\rm 1}\equalcontrib,
    Liangjing Shao\textsuperscript{\rm 1,\rm 2},
    Zihang Yu\textsuperscript{\rm 1},
    Ziyu Wang\textsuperscript{\rm 6},
    Yin Wang\textsuperscript{\rm 5},
    Jiaxi Wang\textsuperscript{\rm 4},
    Jeremy Yuen-Chun Teoh\textsuperscript{\rm 1},
    Zhiyong Xiong\textsuperscript{\rm 3},
    Huxin Gao\textsuperscript{\rm 1},
    Hongliang Ren\textsuperscript{\rm 1,\rm 2}\corresponding
}
\affiliations{
    \textsuperscript{\rm 1}CUHK,
    \textsuperscript{\rm 2}SLAI,
    \textsuperscript{\rm 3}SYSU,
    \textsuperscript{\rm 4}Durham,
    \textsuperscript{\rm 5}NYU,
    \textsuperscript{\rm 6}WFU
}

\begin{document}

\maketitle

\begin{abstract}
Autonomous endoscopic navigation can reduce clinicians' operational burden, yet robust control remains challenging due to tissue deformation, transient occlusions, and rapidly changing viewpoints. Existing learning-based policies typically predict actions from current observations without explicitly modeling future anatomical dynamics, limiting their robustness and reliability in safety-critical settings. World Action Models (WAMs) offer a promising alternative by coupling predictive visual dynamics with action generation, but extending them to robotic endoscopy remains challenging due to limited training data, restricted viewpoint diversity, deformable anatomy, and high inference latency. We present \textbf{EndoWAM}, which is, to our knowledge, the first WAM for generalizable robotic endoscopic navigation. EndoWAM introduces future grounding, which predicts task-relevant target regions in future observations from intermediate denoising features of a video world model. Specifically, EndoWAM couples a lightweight diffusion transformer for future target-region prediction with a discrete action expert through a shared predictive representation. This design injects target-aware supervision into predictive dynamics modeling, improving robustness to visual degradation and viewpoint changes while enabling real-time control in a single denoising pass. We further introduce \textbf{EndoMotion}, a robotic endoscopic motion dataset spanning three anatomically distinct procedures: ureteroscopy, esophagoscopy, and endoscopic retrograde cholangiopancreatography (ERCP). EndoWAM consistently outperforms all baselines and alternative grounding strategies, while demonstrating strong zero-shot generalization to unseen viewpoints, environments, and targets. These results establish EndoWAM as a predictive, target-grounded framework for accurate, generalizable, and long-horizon navigation in deformable and visually constrained endoscopic environments.
\end{abstract}


\section{1 Introduction}

Autonomous endoscopic navigation can substantially alleviate clinicians' operational burden during interventional procedures. Achieving robust and consistent navigation is essential across diverse endoscopic applications, including ureteroscopy, esophagoscopy, and endoscopic retrograde cholangiopancreatography (ERCP)~\cite{Alian2023FlexibleEndoscopy,ercp2}. However, tissue deformation during intervention induces complex and rapidly evolving visual dynamics, while transient occlusions further degrade visual observations, posing significant challenges to vision-based navigation methods. Recent learning-based approaches, including imitation learning (IL) and vision-language-action (VLA) models, have been investigated for specific surgical and endoscopic tasks~\cite{SRT_H_2025,bilivla2026}. Nevertheless, these approaches typically predict actions from current observations without explicitly modeling future environment dynamics. This limited predictive capability hinders their deployment in safety-critical surgical scenarios, where reliable and predictable behavior is essential.

\begin{figure}[t]
\centering
\includegraphics[width=\columnwidth]{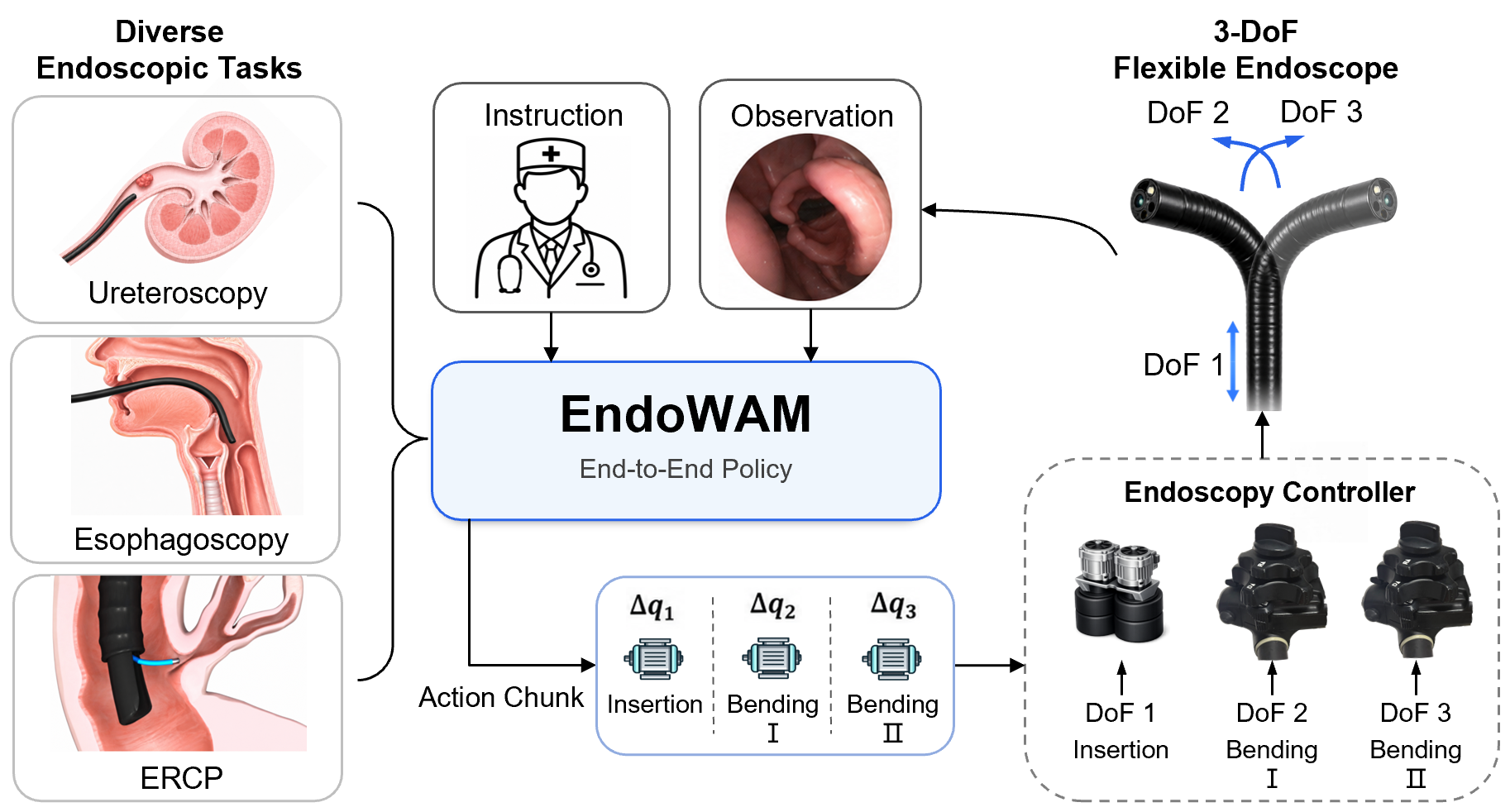}
\caption{\textbf{Overview of EndoWAM for autonomous navigation across diverse endoscopic tasks.} The policy maps the current observation and instruction to action chunks that control the three degrees of freedom of a flexible endoscope.}
\label{fig:endowam-overview}
\end{figure}

World Action Models (WAMs) have recently emerged as a promising paradigm for robotic manipulation and visuomotor control~\cite{li2026causal,li2025uva,bi2025motus}. Unlike VLA models that directly map observations to actions, WAMs incorporate inverse-dynamics objectives or jointly model visual observations and action distributions, enabling them to capture physical dynamics and task-relevant temporal structure~\cite{kim2026cosmospolicy,ma2026dit4dit,ye2026gigaworld}. Extending WAMs to robotic endoscopy, however, presents several challenges. First, ethical constraints and the high costs of clinical procedures limit the collection of large-scale navigation data. Second, confined anatomical workspaces restrict viewpoint diversity, providing sparse supervision for learning visual dynamics. Third, deformable anatomy and transient occlusions complicate target-directed navigation and increase the risk of unsafe tissue interactions. Finally, the high inference latency of existing WAMs limits their responsiveness to rapidly changing endoscopic observations, hindering deployment in time-sensitive surgical scenarios.

We address these challenges with \textbf{EndoWAM}, which is, to our knowledge, the first World Action Model for generalizable robotic endoscopic navigation, as illustrated in Fig.~\ref{fig:endowam-overview}. EndoWAM introduces a future grounding paradigm that uses latent-space supervision from task-relevant regions in future observations to shape predictive visual representations. Specifically, a diffusion transformer (DiT), conditioned on intermediate denoising features from a video world model, predicts future target-region latents. Unlike previous grounding methods that localize targets only in the current observation~\cite{roboground,lisa,vip,khanam2024yolov11,ecot,graspvla,reconvla}, future grounding captures the spatial evolution of task-relevant targets. This design promotes fine-grained, target-aware temporal representations, enabling the action policy to exploit predictive visual dynamics for robust and precise endoscopic navigation.

To alleviate the scarcity of temporally aligned endoscopic navigation data, we further introduce \textbf{EndoMotion}, a multi-procedure robotic endoscopy dataset comprising long-horizon trajectories across ureteroscopy, esophagoscopy, and ERCP. We collect human teleoperation trajectories in corresponding anatomical phantoms and use a fine-tuned Grounding DINO~\cite{liu2024groundingdino} to densely annotate task-relevant target regions. These annotations provide target-relative action labels and future target regions for grounding supervision. We additionally introduce viewpoint augmentation by jointly rotating each trajectory and its target annotations across multiple roll angles, substantially increasing visual diversity and viewpoint coverage.

Experiments across diverse long-horizon robotic endoscopic navigation tasks demonstrate that our future grounding paradigm consistently outperforms existing visual grounding strategies. Attention visualizations further show that EndoWAM focuses on task-relevant target regions, enabling precise and temporally coherent navigation. Ablation studies confirm that viewpoint augmentation substantially improves generalization to unseen viewpoints. Moreover, EndoWAM exhibits strong zero-shot generalization to unseen environments and targets. Collectively, these results establish EndoWAM as a robust, target-aware, and generalizable framework for long-horizon navigation in complex and visually constrained endoscopic environments.




In summary, our main contributions are as follows:
\begin{itemize}
    \item We propose \textbf{EndoWAM}, which is, to our knowledge, the \textbf{first} World Action Model for generalizable robotic endoscopic navigation. By conditioning the policy on intermediate denoising features of a video world model, EndoWAM incorporates predictive visual dynamics while enabling real-time navigation.

    \item We introduce a future grounding paradigm that predicts task-relevant regions in future latent states, enabling fine-grained, target-aware temporal representation learning for precise endoscopic navigation.

    \item We construct \textbf{EndoMotion}, a robotic endoscopy motion dataset spanning ureteroscopy, esophagoscopy, and ERCP, with dense target annotations, action labels, and viewpoint augmentation.

    \item Extensive real-world experiments demonstrate state-of-the-art performance, validate the proposed grounding strategies, and show that the policy generalizes effectively to unseen viewpoints, environments, and targets.
\end{itemize}

\section{2 Related Work}

\noindent\textbf{Vision-Language-Action and World Action Models.}
Learning-based visuomotor control has evolved from imitation policies~\cite{chi2023diffusion,zhao2023act} to VLAs~\cite{OpenVLA_2024,black2024pi_0,intelligence2025pi_}. Both learn action policies from demonstrations, while VLAs additionally inherit semantic reasoning capabilities from pretrained vision-language models (VLMs). However, these methods typically predict actions from current observations without explicitly modeling how the environment evolves under robot actions. More recently, World Action Models (WAMs)~\cite{li2026causal,cen2025worldvla,li2025uva,bi2025motus,kim2026cosmospolicy,ma2026dit4dit,ye2026gigaworld} have incorporated inverse-dynamics objectives or jointly modeled visual observations and action distributions, often building on large-scale video world models~\cite{agarwal2025cosmos,assran2025vjepa2}. Their predictive representations capture physical dynamics and task-relevant temporal structure beyond the current observation. While existing WAMs primarily target robotic manipulation, EndoWAM extends predictive visuomotor modeling to deformable and partially observable anatomical environments and grounds future visual dynamics in task-relevant target regions.

\begin{figure*}[t]
\centering
\includegraphics[width=0.9\textwidth]{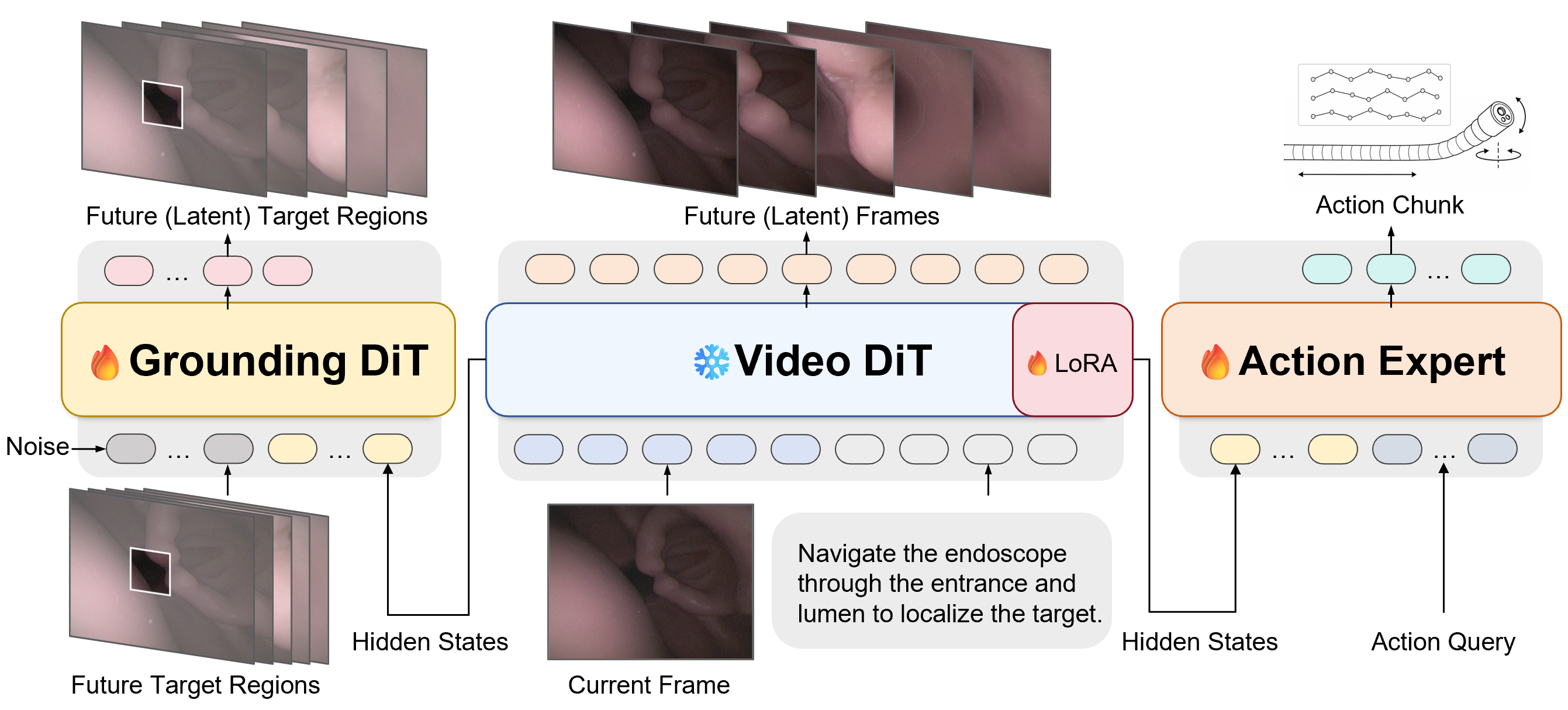}
\caption{\textbf{EndoWAM architecture.} The Video DiT models future latent dynamics from the current frame and instruction. Its hidden states condition the Grounding DiT to reconstruct future target regions and the action expert to generate an action chunk.}
\label{fig:endowam-architecture}
\end{figure*}

\noindent\textbf{Robot Learning for Autonomous Surgical Robotics.}
Surgical robotics is increasingly transitioning from rule-based control to data-driven policies capable of interacting with complex and deformable anatomy~\cite{schmidgall2024foundation,long2025surgical,SRT_H_2025}. Most existing approaches rely on imitation learning~\cite{kim2024srt,moghani2025sufia}, while reinforcement learning has also been explored under simulated supervision~\cite{singh2023retraction}. In flexible endoscopy, autonomous navigation is typically addressed through visual servoing or platform-specific reinforcement learning~\cite{endoscopic_control2,britle1,zhang2024ai}. More recently, VLA-based methods have been investigated for scrub-nurse support~\cite{li2025robonurse}, instrument tracking~\cite{EndoVLA_2025}, and biliary navigation~\cite{bilivla2026}. However, these methods primarily infer actions from current observations without explicitly modeling future anatomical dynamics. In contrast, EndoWAM conditions action generation on predictive intraluminal representations and, to our knowledge, is the first World Action Model for generalizable robotic endoscopic navigation.

\noindent\textbf{Visual Grounding Methods for Robotic Control.}
Fine-grained robotic control requires policies to identify and attend to task-relevant visual regions. Existing grounding methods either rely on external perception models or learn grounding jointly with the control policy. RoboGround~\cite{roboground} uses LISA~\cite{lisa} to segment target and background regions as auxiliary inputs, while VIP~\cite{vip} crops and enlarges target regions detected by YOLOv11~\cite{khanam2024yolov11}; both therefore depend on the accuracy of external grounding models. Alternatively, ECoT~\cite{ecot} and GraspVLA~\cite{graspvla} predict bounding boxes before action generation, whereas ReconVLA~\cite{reconvla} predicts gaze regions from policy representations to provide implicit grounding supervision. Despite their different formulations, these approaches ground task-relevant regions only in the current observation. In contrast, EndoWAM introduces future grounding, which predicts task-relevant target regions in future observations from intermediate world model representations, thereby injecting target-centric supervision into predictive visual dynamics.

\section{3 Method}

\subsection{3.1 Problem Formulation}

Unlike conventional VLA policies that directly map current visual observations $o_t$ and language instructions $l$ to actions $a_t$, EndoWAM adopts a \emph{predict-latent-dynamics-then-ground-and-act} paradigm. Given $o_t$ and $l$, the video world model predicts a latent representation of future visual dynamics $\mathbf{z}_{t+1}$, whose intermediate denoising features jointly support multi-step action generation and future target grounding:
\begin{equation}
\begin{gathered}
\mathbf{z}_{t+1}
\sim p_v\!\left(\cdot \mid o_t,l\right),\\
\mathbf{a}_t
\sim p_a\!\left(
\cdot \mid \mathcal{H}\!\left(\mathbf{z}_{t+1}^{\tau_v}\right)
\right),\\
\mathbf{r}_{t+1}
\sim p_r\!\left(
\cdot \mid \mathcal{H}\!\left(\mathbf{z}_{t+1}^{\tau_v}\right)
\right).
\end{gathered}
\label{eq:endo-wam-formulation}
\end{equation}
Here, $\mathbf{z}_{t+1}^{\tau_v}$ denotes the intermediate latent state at flow step $\tau_v$, which progressively approaches $\mathbf{z}_{t+1}$ as $\tau_v \rightarrow 0$. The operator $\mathcal{H}$ extracts hidden representations from the denoising process without explicitly decoding future observations into pixel-space frames.

The variable $\mathbf{r}_{t+1}$ denotes task-relevant target regions in future observations. Training jointly models future visual dynamics, multi-step actions, and target regions through the conditional latent-action-region distribution:
\begin{equation}
\mathbf{z}_{t+1},\mathbf{a}_t,\mathbf{r}_{t+1}
\sim
p_{var}\!\left(\cdot \mid o_t,l\right).
\label{eq:joint-distribution}
\end{equation}

\subsection{3.2 Model Architecture}

\noindent\textbf{Overview.}
EndoWAM is designed for generalizable endoscopic navigation, generating instruction-conditioned three-DoF controls that steer a flexible endoscope toward anatomical targets despite occlusions and specular artifacts. Rather than relying solely on the current observation, EndoWAM grounds its policy in predicted future dynamics. During training, a video world model captures future visual dynamics in a single denoising pass, and the resulting shared representation conditions an action decoder for multi-step control and a Grounding DiT for reconstructing future target-region latents (Fig.~\ref{fig:endowam-architecture}). At inference the world model still performs the same single denoising pass to produce the shared representation, but future-frame decoding and the Grounding DiT branch are dropped, so future-grounded representation learning adds no deployment cost.

\noindent\textbf{Architecture.}
The world model is initialized from Cosmos-Predict2.5-2B~\cite{agarwal2025cosmos}, a pretrained video DiT with a text encoder and video VAE. The instruction is encoded and injected into the DiT tokens through cross-attention, while endoscopic observations are mapped to latent video tokens by the VAE. The current observation is retained as a clean conditioning latent, whereas future latent tokens are perturbed under a rectified-flow schedule and jointly denoised. We apply LoRA~\cite{hu2022lora} to the attention and feed-forward projections to efficiently adapt the pretrained video prior to the endoscopic domain.

To instantiate $\mathcal H$ in Eq.~\ref{eq:endo-wam-formulation}, we extract an intermediate activation after a single denoising pass as the shared representation
$H=\mathcal H(\mathbf z_{t+1}^{\tau_v})$
and flatten it into a sequence of tokens. Two prediction heads are built upon $H$. The action decoder processes learnable action queries through transformer blocks that combine bidirectional self-attention with cross-attention to $H$, followed by an MLP that predicts per-axis categorical logits. The bidirectional attention enables each action query to incorporate the full temporal context of the action chunk. The Grounding DiT reconstructs the VAE latents of target crops in future frames, conditioned on a projected and resampled representation of $H$. This future grounding objective encourages $H$ to capture the spatial evolution of task-relevant targets, yielding predictive features for action decoding.

\subsection{3.3 Training and Inference}

\noindent\textbf{Dataset.}
To provide aligned control and grounding supervision for endoscopic navigation, we construct \textbf{EndoMotion}, comprising $515$ trajectories and $0.98$M frames collected from three procedures: ureteroscopy, esophagoscopy, and ERCP. We use Grounding DINO~\cite{liu2024groundingdino} to localize the instructed anatomical targets and apply Kalman filtering to obtain temporally consistent oriented bounding-box (OBB) tracks. Each track provides two aligned supervision signals: the target-center offset relative to the endoscopic view is converted into discrete action labels, while target regions in future frames are cropped to provide future grounding supervision. To improve robustness to viewpoint variations, we further introduce viewpoint augmentation by rotating each trajectory and its corresponding OBB annotations at $45^\circ$ intervals, expanding the dataset to $4{,}120$ sequences and $7.84$M frames. By aligning observations, target-relative actions, and future target regions across anatomically distinct procedures and viewpoints, EndoMotion supports the joint learning of predictive dynamics, action generation, and future grounding. Per-procedure statistics are reported in Appendix~C, and the full annotation pipeline is detailed in Appendix~B.

\begin{figure}[t]
\centering
\includegraphics[width=\columnwidth]{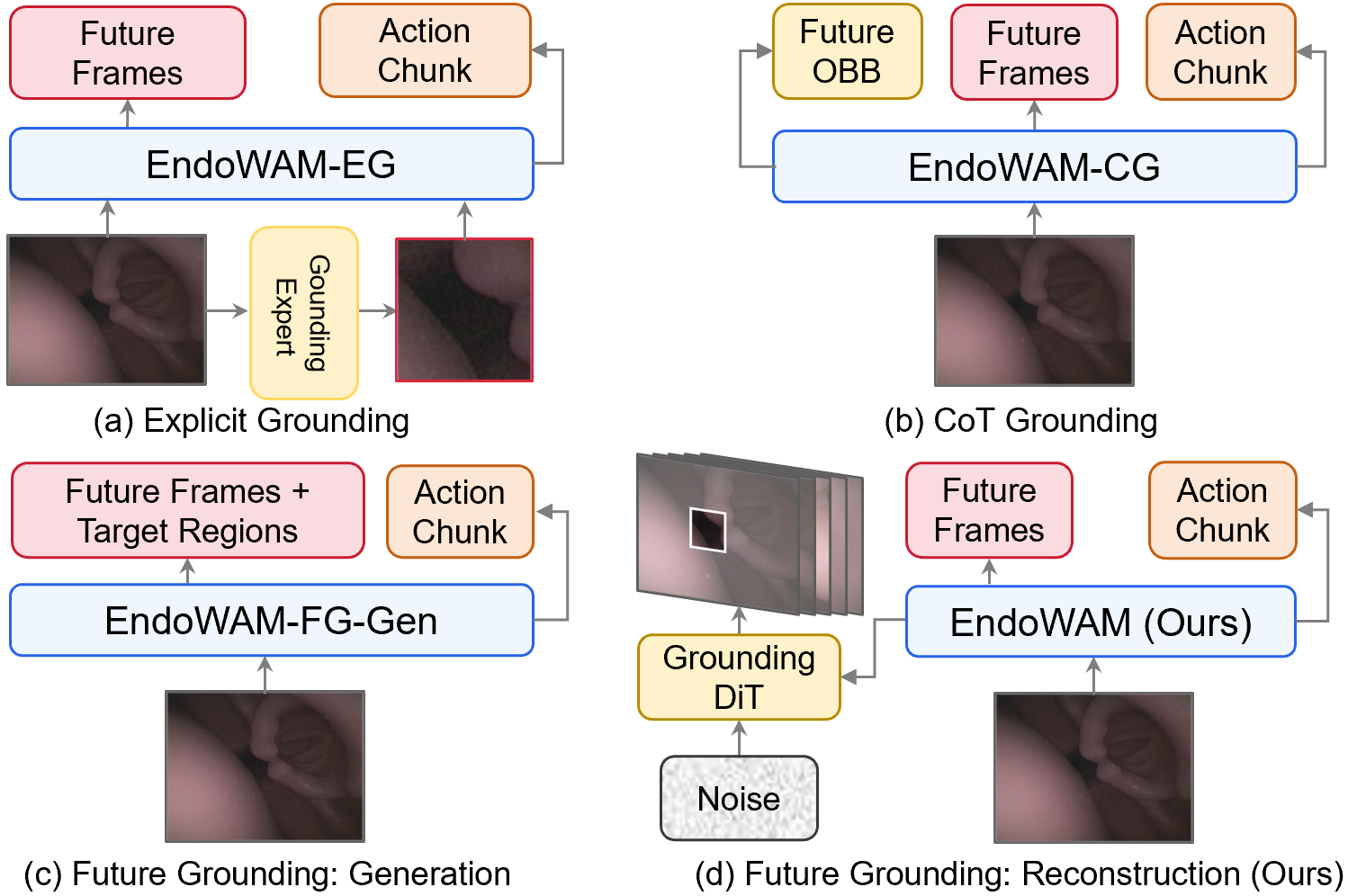}
\caption{\textbf{Conceptual comparison of different paradigms in EndoWAM.}
(a) \textbf{EndoWAM-EG} performs explicit input grounding by injecting a grounding-expert crop of the current target. 
(b) \textbf{EndoWAM-CG} performs CoT grounding by regressing future OBBs from the shared predictive representation.
(c) \textbf{EndoWAM-FG-Gen} jointly generates future frames and target regions through the world model's generative pathway.
(d) \textbf{EndoWAM (Ours)} performs future grounding by reconstructing the latent target regions of future frames with a Grounding DiT.}
\label{fig:grounding-variants}
\end{figure}

\noindent\textbf{Training objective.}
We jointly optimize EndoWAM for predictive visual dynamics, multi-step action generation, and future target grounding, using three corresponding objectives. For video co-training, we sample $\varepsilon\sim\mathcal N(0,I)$ and a flow time $\tau$, interpolate the future latents as $x_\tau=(1-\tau)x_0+\tau\varepsilon$, and regress the velocity field over future positions:
\begin{equation}
\mathcal L_{\text{video}}
=
\mathbb E_{\tau,\varepsilon}
\Big[
\big\|
v_\theta(x_\tau,\tau \mid o_t,l)
-
(\varepsilon-x_0)
\big\|_2^2
\Big].
\end{equation}
Here, $\tau$ is resampled for each training example, whereas $\tau_v$ in Eq.~\ref{eq:endo-wam-formulation} denotes the flow time at which the shared representation $H$ is extracted.

The action decoder is trained with per-axis weighted cross-entropy under a validity mask $m$:
\begin{equation}
\mathcal L_{\text{action}}
=
\frac{
\sum_{k,a}
m_{k,a}\,
\mathrm{CE}
\big(
p_\theta(\cdot\mid q_k,H)_a,
y_{k,a};
w_a
\big)
}{
\sum_{k,a}m_{k,a}
},
\end{equation}
where $k$ indexes the action chunk, $a$ indexes the three motion axes, $y_{k,a}$ is the target action bin, and $w_a$ denotes the corresponding class weights.

For future grounding, the Grounding DiT denoises the target-region latent $\mathbf{z}^{\mathrm{tar}}_n$ of future frame $n$ at diffusion step $s$ under a cosine schedule:
\begin{equation}
\label{eq:future-grounding}
\mathcal L_{\text{ground}}
=
\mathbb E_{n,s,\varepsilon}\!
\Big[
u_n
\big\|
\varepsilon-
\varepsilon_\theta\!
\big(
\sqrt{\bar\alpha_s}\mathbf{z}^{\mathrm{tar}}_n
+
\sqrt{1-\bar\alpha_s}\varepsilon,
s,H
\big)
\big\|_2^2
\Big],
\end{equation}
where the target mask $u_n\in\{0,1\}$ restricts supervision to future frames with detected targets, so missed detections contribute no gradient. The overall objective is
\begin{equation}
\mathcal L
=
\mathcal L_{\text{action}}
+
\lambda_v\mathcal L_{\text{video}}
+
\lambda_r\mathcal L_{\text{ground}},
\end{equation}
where $\lambda_v$ and $\lambda_r$ balance action learning, video co-training, and future grounding.

\begin{figure*}[t]
\centering
\includegraphics[width=0.7\textwidth]{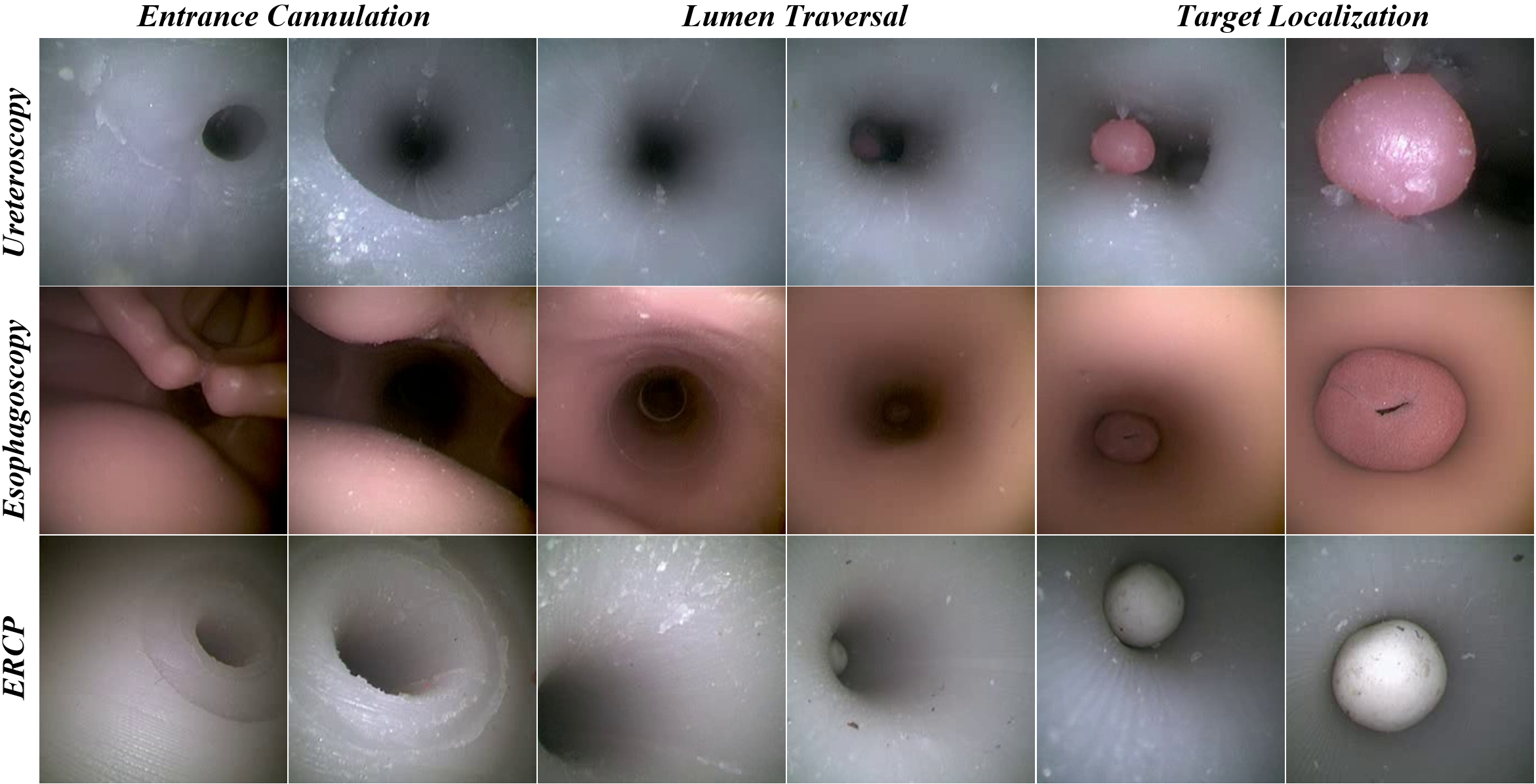}
\caption{\textbf{Closed-loop deployment trajectories.} Representative EndoWAM rollouts across ureteroscopy, esophagoscopy, and ERCP, covering the three navigation phases: entrance cannulation, lumen traversal, and target localization.}
\label{fig:deployment-trajectories}
\end{figure*}

\subsection{3.4 Controlled Variants for Grounding WAM}

To investigate how visual attention allocation affects grounding in WAMs, we construct a set of controlled variants inspired by representative grounding strategies in VLA models, as grounding in WAMs remains underexplored. For a fair comparison, all variants share the same backbone, tokenization scheme, and training protocol whenever applicable.

As illustrated in Fig.~\ref{fig:grounding-variants}, EndoWAM-EG performs explicit input grounding by augmenting the current observation with a cropped and resized target region produced by a fine-tuned grounding expert. EndoWAM-CG directly regresses future OBB parameters $(c_x,c_y,w,h,\sin 2\theta,\cos 2\theta)$ from the shared representation $H$. EndoWAM-FG-Gen instead incorporates future target regions into the world model's generation target, allowing grounding to be learned through the generative pathway rather than a separate reconstruction branch. These variants represent input-level, regression-based, and generative grounding strategies, respectively, and are evaluated under matched settings in Sec.~4.3.

\section{4 Experiments}

We design our experiments around the following questions:
\begin{itemize}
    \item Can EndoWAM outperform state-of-the-art methods on long-horizon endoscopic navigation under unseen viewpoints? (Sec.~4.2)
    \item Does our \textbf{future grounding} approach outperform alternative visual grounding paradigms? (Sec.~4.3)
    \item How do the grounding paradigm and other key design choices affect the overall performance of EndoWAM? (Sec.~4.4)
    \item Can EndoWAM generalize to unseen environments and unseen targets? (Sec.~4.5)
    \item Is EndoWAM efficient enough for closed-loop endoscopic control compared with existing methods? (Sec.~4.6)
\end{itemize}

\begin{table*}[t]
\centering
\small
\setlength{\tabcolsep}{6pt}
\begin{tabular}{@{}lccccccc@{}}
\toprule
Method &
\begin{tabular}[c]{@{}c@{}}Discrete\\Action\end{tabular} &
\begin{tabular}[c]{@{}c@{}}Visual\\Grounding\end{tabular} &
\begin{tabular}[c]{@{}c@{}}Dynamics\\Prior\end{tabular} &
Ureteroscopy &
Esophagoscopy &
ERCP &
\textbf{Average} \\
\midrule
Diffusion Policy~\cite{chi2023diffusion} & \xmark & \xmark & \xmark & 3.1 & 0.0 & 3.1 & 2.1 \\
Qwen3DiT & \xmark & \xmark & \xmark & 9.4 & 3.1 & 12.5 & 8.3 \\
$\pi_{0.5}$~\cite{intelligence2025pi_} & \xmark & \xmark & \xmark & 18.8 & 6.3 & 21.9 & 15.6 \\
Qwen3AE & \cmark & \xmark & \xmark & 15.6 & 12.5 & 25.0 & 17.7 \\
GR00T-N1.7~\cite{bjorck2025gr00t} & \xmark & \xmark & \xmark & 25.0 & 21.9 & 34.4 & 27.1 \\
\textbf{EndoWAM (Ours)} & \cmark & \cmark & \cmark & \textbf{90.6} & \textbf{56.3} & \textbf{93.8} & \textbf{80.2} \\
\midrule
\multicolumn{8}{@{}l}{\textit{EndoWAM variants}} \\
\midrule
EndoWAM-CG  & \cmark & \cmark & \cmark & 46.9 & 37.5 & 53.1 & 45.8 \\
EndoWAM-FG-Gen  & \cmark & \cmark & \cmark & 43.8 & 46.9 & 62.5 & 51.0 \\
EndoWAM-EG & \cmark & \cmark & \cmark & 71.9 & 43.8 & 68.8 & 61.5 \\
\textbf{EndoWAM (Ours)} & \cmark & \cmark & \cmark & \textbf{90.6} & \textbf{56.3} & \textbf{93.8} & \textbf{80.2} \\
\bottomrule
\end{tabular}
\caption{\textbf{Closed-loop navigation success rates (\%) across three endoscopic procedures.}
For each procedure, $32$ trials are initialized from a randomly sampled viewpoint and evaluated at uniformly spaced roll offsets spanning the full $360^\circ$ range, with test orientations offset from the eight roll angles used for training augmentation.}
\label{tab:endowam-results}
\end{table*}

\begin{table}[t]
\centering
\small
\setlength{\tabcolsep}{3pt}
\begin{tabular}{@{}lcccc@{}}
\toprule
Variant & Ureteroscopy & Esophagoscopy & ERCP & \textbf{Avg.} \\
\midrule
w/o Grounding DiT & 37.5 & 28.1 & 46.9 & 37.5 \\
w/o viewpoint aug. & 31.3 & 15.6 & 37.5 & 28.1 \\
\textbf{Full model} &
\textbf{90.6} & \textbf{56.3} & \textbf{93.8} & \textbf{80.2} \\
\bottomrule
\end{tabular}
\caption{\textbf{Ablation of future grounding and viewpoint augmentation.}
Closed-loop navigation success rates (\%) across three procedures.}
\label{tab:endowam-ablation}
\end{table}

\begin{table*}[t]
\centering
\small
\setlength{\tabcolsep}{2.5pt}
\begin{tabular}{@{}lccccccc@{}}
\toprule
& \multicolumn{3}{c}{Cross-environment} &
\multicolumn{3}{c}{Cross-target} & \\
\cmidrule(lr){2-4}\cmidrule(lr){5-7}
Method &
Reflectance Shift &
Geometry Shift &
Texture Shift &
Shape Shift &
Size Shift &
Color Shift &
\textbf{Average} \\
\midrule
Qwen3AE & 31.3 & 28.1 & 40.6 & 28.1 & 53.1 & 37.5 & 36.5 \\
GR00T-N1.7~\cite{bjorck2025gr00t} & 53.1 & 37.5 & 53.1 & 43.8 & 59.4 & 34.4 & 46.9 \\
\textbf{EndoWAM (Ours)} & \textbf{93.8} & \textbf{87.5} & \textbf{100.0} & \textbf{84.4} & \textbf{93.8} & \textbf{81.3} & \textbf{90.1} \\
\bottomrule
\end{tabular}
\caption{\textbf{Zero-shot generalization to unseen environments and targets.}
Success rates (\%) over $32$ trials per condition for lumen traversal and target localization; results are not directly comparable to the three-stage evaluation in Table~\ref{tab:endowam-results}.}
\label{tab:zero-shot-generalization}
\end{table*}

\subsection{4.1 Implementation Details}

\noindent\textbf{Deployment and evaluation protocol.}
All policies are deployed on the same three-DoF robotic endoscope and evaluated on physical phantoms for ureteroscopy, esophagoscopy, and ERCP. A trial is considered successful if the endoscope reaches the instructed target and maintains it within the field of view until the rollout terminates; failures are categorized as lumen loss, wall impaction, or timeout. Each policy is evaluated over $32$ trials per procedure. For each procedure, we randomly sample an initial viewpoint and apply uniformly spaced roll offsets spanning the full $360^\circ$ range. These test viewpoints are deliberately unaligned with the eight angles used for viewpoint augmentation, ensuring that every trial evaluates generalization to an unseen viewpoint. Representative EndoWAM rollouts are shown in Fig.~\ref{fig:deployment-trajectories}, with additional demonstrations provided in Appendix~F. Further details on the robotic platform, anatomical phantoms, and success criteria are provided in Appendix~A.

\noindent\textbf{Baselines.}
We compare EndoWAM against five baselines. Diffusion Policy~\cite{chi2023diffusion} serves as a non-VLA visuomotor policy, while $\pi_{0.5}$~\cite{intelligence2025pi_} and GR00T-N1.7~\cite{bjorck2025gr00t} represent large-scale pretrained VLA policies. Qwen3DiT and Qwen3AE share the same Qwen3-VL-2B~\cite{bai2025qwen3} backbone but differ in their action heads: the former adopts a diffusion transformer (DiT) action head, whereas the latter uses our discretized action expert (AE). This controlled comparison isolates the effect of action parameterization under a shared visual prior. All baselines receive identical observations and instructions and are trained on the same rotation-augmented EndoMotion dataset for a matched number of optimization steps. Detailed training configurations for each baseline are provided in Appendix~E.

\noindent\textbf{Model and optimization.}
EndoWAM adapts Cosmos-Predict2.5-2B~\cite{agarwal2025cosmos} using LoRA~\cite{hu2022lora} on the attention and feed-forward projections, while keeping the text encoder and video VAE frozen. The shared representation $H$ is extracted from the block-$17$ activation after a single denoising pass and jointly conditions both the action decoder and the Grounding DiT. Architectural and optimization details are provided in Appendix~D.

\subsection{4.2 Comparison with State-of-the-Art}
We compare EndoWAM with representative imitation-learning and VLA policies under the closed-loop protocol described in Sec.~4.1. To evaluate generalization to unseen viewpoints, all methods are tested over uniformly spaced roll offsets spanning the full $360^\circ$ range, with test viewpoints unaligned with those used for training augmentation. The comparison spans action-generation mechanisms and examines the contributions of discrete action prediction, visual grounding, and predictive dynamics modeling.

\noindent\textbf{Results.}
The upper block of Table~\ref{tab:endowam-results} summarizes the comparison. EndoWAM succeeds in $77$ of $96$ trials, achieving an average success rate of $80.2\%$ and outperforming the strongest baseline, GR00T-N1.7 ($27.1\%$), by $53.1$ percentage points. It also achieves the highest success rate across all three procedures, with $90.6\%$ on ureteroscopy, $56.3\%$ on esophagoscopy, and $93.8\%$ on ERCP. A breakdown of the observed failure modes is reported in Appendix~G. These results demonstrate that predictive dynamics priors and our future grounding approach consistently benefit diverse endoscopic navigation tasks, while enabling strong generalization to unseen viewpoints. Moreover, Qwen3AE outperforms Qwen3DiT under the same vision-language backbone, validating the effectiveness of our discrete action expert over the diffusion-based action head in this setting.

\subsection{4.3 Comparison with EndoWAM Variants}
We evaluate the controlled grounding variants introduced in Fig.~\ref{fig:grounding-variants} and Sec.~3.4 under the same closed-loop protocol and unseen-viewpoint setting as in Sec.~4.2. All variants share matched training and evaluation configurations, with results reported in the lower block of Table~\ref{tab:endowam-results}.

\noindent\textbf{Results.}
EndoWAM achieves the highest average success rate of $80.2\%$, outperforming the strongest variant, EndoWAM-EG, by $18.7$ points, with gains of $18.7$, $12.5$, and $25.0$ points on ureteroscopy, esophagoscopy, and ERCP, respectively. Directly regressing future OBBs in EndoWAM-CG yields an average success rate of $45.8\%$, while incorporating future target regions into the generative objective improves performance to $51.0\%$. Explicit input grounding further increases the average success rate to $61.5\%$. Nevertheless, all variants remain substantially below our reconstruction-based future grounding, indicating that reconstructing future target regions from predictive denoising features more effectively couples visual dynamics, target-aware attention, and multi-step action generation.

\begin{figure}[t]
\centering
\includegraphics[width=0.8\columnwidth]{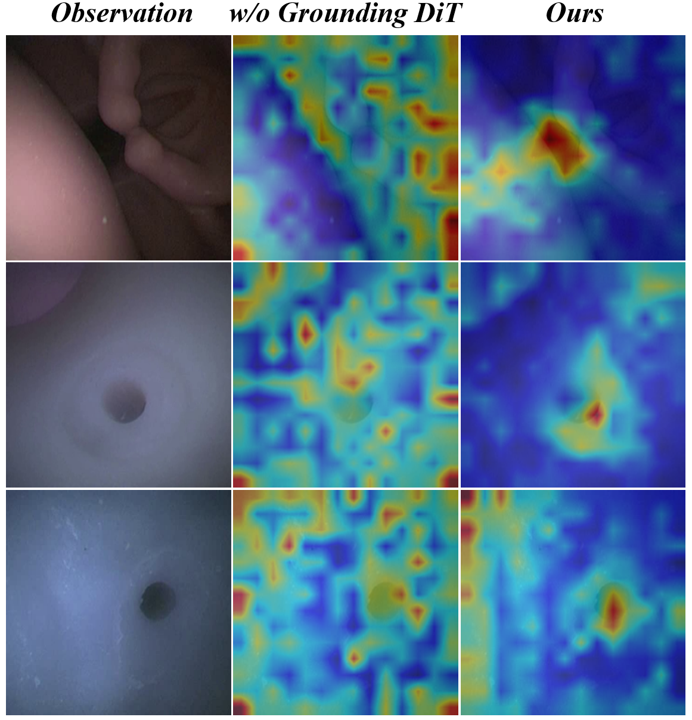}
\caption{\textbf{Attention visualization for Grounding DiT ablation.} Without the Grounding DiT, the policy attends diffusely to specular highlights and mucosal textures, whereas the full model concentrates its attention around the task-relevant target even under darkness and partial occlusion.}
\label{fig:target-visualization}
\end{figure}

\subsection{4.4 Ablation Study}

Table~\ref{tab:endowam-ablation} evaluates the Grounding DiT and viewpoint augmentation through independent single-factor ablations. Removing the Grounding DiT reduces the average success rate from $80.2\%$ to $37.5\%$, demonstrating the importance of future target supervision for reliable navigation. Removing viewpoint augmentation lowers the average success rate to $28.1\%$, indicating that exposure to diverse viewpoints is critical for generalization to unseen orientations. Figure~\ref{fig:target-visualization} provides qualitative evidence: without the Grounding DiT, policy attention is diffuse and frequently captured by specular highlights and mucosal textures, whereas the full model concentrates its attention around the task-relevant target even under darkness and partial occlusion. These results show that the Grounding DiT promotes target-aware predictive representations, while viewpoint augmentation improves viewpoint robustness.

\subsection{4.5 Generalization Capability}

\begin{figure}[t]
\centering
\includegraphics[width=0.9\columnwidth]{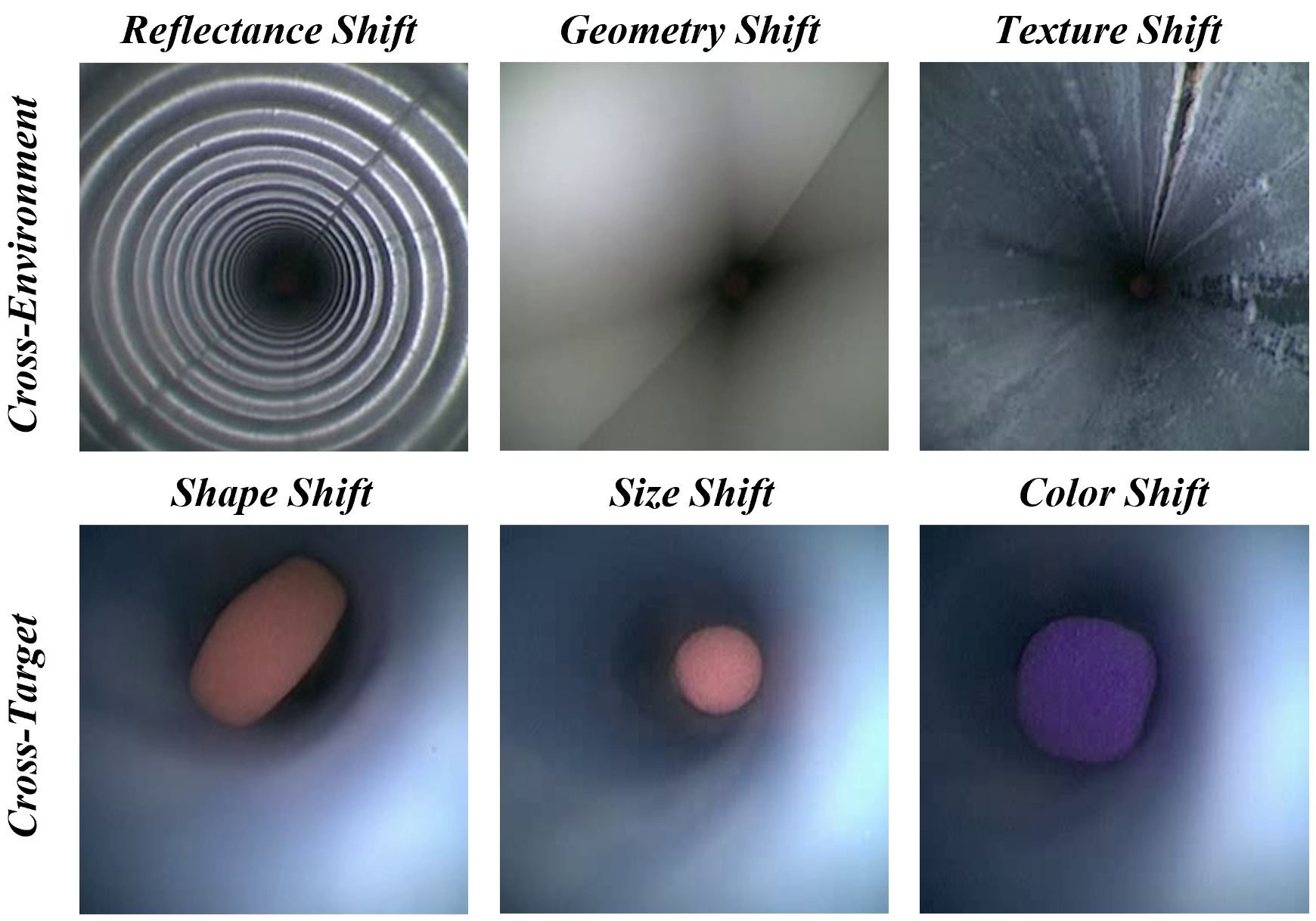}
\caption{\textbf{Zero-shot deployment examples.} Representative observations from EndoWAM rollouts under unseen cross-environment shifts in lumen reflectance, geometry, and texture, and cross-target shifts in target shape, size, and color.}
\label{fig:zero-shot-settings}
\end{figure}

\noindent\textbf{Evaluation setup.}
We evaluate zero-shot generalization across six conditions unseen during training. As shown in Fig.~\ref{fig:zero-shot-settings}, cross-environment shifts alter lumen reflectance, geometry, and texture, whereas cross-target shifts alter target shape, size, and color. Each condition comprises $32$ closed-loop trials following a two-stage protocol of lumen traversal and target localization.

\noindent\textbf{Results.}
As reported in Table~\ref{tab:zero-shot-generalization}, EndoWAM achieves an average success rate of $90.1\%$, outperforming the strongest baseline, GR00T-N1.7, by $43.2$ percentage points. EndoWAM maintains consistently high performance across all conditions, reaching $100.0\%$ under texture shift and exceeding $80\%$ across all target variations. These results demonstrate that jointly modeling visual dynamics and target-aware grounding enables robust transfer to unseen lumen appearances and target configurations.

\begin{table}[t]
\centering
\small
\setlength{\tabcolsep}{3pt}
\begin{tabular}{@{}lccc@{}}
\toprule
Policy & \begin{tabular}[c]{@{}c@{}}Dynamics\\Prior\end{tabular} & Params. & \begin{tabular}[c]{@{}c@{}}Control\\Freq. (Hz)\end{tabular} \\
\midrule
Qwen3AE & \xmark & 2.2B & 12.1 \\
GR00T-N1.7 & \xmark & 1.6B & 8.9 \\
Cosmos Policy~\cite{kim2026cosmospolicy} & \cmark & 2.0B & 1.1 \\
\textbf{EndoWAM (Ours)} & \cmark & 2.1B & \textbf{7.5} \\
\bottomrule
\end{tabular}
\caption{\textbf{Closed-loop deployment efficiency with a single NVIDIA RTX 5090.}}
\label{tab:deployment-efficiency}
\end{table}

\subsection{4.6 Deployment Efficiency}
Table~\ref{tab:deployment-efficiency} compares closed-loop inference efficiency on a single NVIDIA RTX 5090. At a comparable parameter scale, EndoWAM achieves a competitive control frequency of $7.5$\,Hz relative to VLA policies, while retaining an explicit dynamics prior. Moreover, EndoWAM is $6.8\times$ faster than Cosmos Policy ($1.1$\,Hz), because it predicts an action chunk from intermediate features obtained in a single denoising pass, whereas Cosmos Policy must iteratively denoise future video representations before generating actions. Combined with the improved navigation success rates reported in Table~\ref{tab:endowam-results}, these results demonstrate that EndoWAM enhances navigation accuracy while meeting the real-time control requirements of endoscopic navigation.

\section{5 Conclusion}

We presented \textbf{EndoWAM}, which is, to our knowledge, the first World Action Model for generalizable robotic endoscopic navigation. EndoWAM introduces future grounding, which predicts task-relevant regions in future observations to inject target-aware supervision into intermediate world-model representations. This design enables the policy to capture future visual dynamics and target evolution, improving multi-step navigation under occlusions, visual degradation, and viewpoint changes. We further constructed \textbf{EndoMotion}, a multi-procedure robotic endoscopy dataset with aligned visual observations, action labels, future target annotations, and viewpoint augmentation. Closed-loop experiments on physical phantoms for ureteroscopy, esophagoscopy, and ERCP show that EndoWAM achieves an average success rate of $80.2\%$, outperforming the strongest baseline by $53.1$ percentage points. It also attains a $90.1\%$ average zero-shot success rate under unseen environment and target shifts. By generating action chunks from shared predictive representations in a single denoising pass, EndoWAM operates at $7.5$\,Hz, demonstrating a strong balance among navigation accuracy, generalization, and real-time control.

\bibliography{bibliography}

%


%

\end{document}